\documentclass[twoside]{article}
\usepackage{PRIMEarxiv}

\usepackage{times}
\usepackage{soul}
\usepackage{url}
\usepackage[hidelinks]{hyperref}
\usepackage[utf8]{inputenc}
\usepackage[small]{caption}
\usepackage{graphicx}
\usepackage{subcaption}
\usepackage{amsmath}
\usepackage{amsthm}
\usepackage{booktabs}
\usepackage{algorithm}
\usepackage{algorithmic}
\usepackage[switch]{lineno}
\usepackage{amssymb}
\usepackage{booktabs}
\usepackage[dvipsnames]{xcolor}
\usepackage{svg}

\newtheorem{definition}{Definition}

\title{Markov Constraint as Large Language Model Surrogate}

\author{
Alexandre Bonlarron \\
Université Côte d’Azur, Inria, France \\
Université Côte d’Azur, CNRS, I3S, France \\
alexandre.bonlarron@gmail.com
\and
Jean-Charles Régin \\
    Université Côte d’Azur, CNRS, I3S, France\\
    jcregin@gmail.com
}

\begin{document}
 \maketitle
\begin{abstract}
   This paper presents \emph{NgramMarkov}, a variant of the Markov constraints. It is dedicated to text generation in constraint programming (CP).
It involves a set of n-grams (i.e., sequence of $n$ words) associated with probabilities given by a large language model (LLM). It limits the product of the probabilities of the n-gram of a sentence. 

The propagator of this constraint can be seen as an extension of the \emph{ElementaryMarkov} constraint propagator, incorporating the LLM distribution instead of the maximum likelihood estimation of n-grams. It uses a gliding threshold, i.e., it rejects n-grams whose local probabilities are too low, to guarantee balanced solutions.
It can also be combined with a "look-ahead" approach to remove n-grams that are very unlikely to lead to acceptable sentences for a fixed-length horizon. This idea is based on the \emph{MDDMarkovProcess} constraint propagator, but without explicitly using an MDD (Multi-Valued Decision Diagram).

The experimental results show that the generated text is valued in a similar way to the LLM perplexity function.
Using this new constraint dramatically reduces the number of candidate sentences produced, improves computation times, and allows larger corpora or smaller n-grams to be used. A real-world problem has been solved for the first time using 4-grams instead of 5-grams.

\end{abstract}

\section{Introduction}
Language modeling is a fundamental challenge in natural language processing (NLP). Language models (LMs), such as n-grams \cite{shannon-ngram:51}, recurrent neural networks (RNNs), long short-term memory (LSTM)  networks \cite{LSTM:1997}, and transformers \cite{vasmani-et-al:2017} (LLM), are tools employed to model natural language. Which is to say, computing the probability that a sequence of words belongs to a language. In this paper, LMs are looked at through the lens of a constrained text generation task (i.e., producing an output that satisfies a set of requirements). In this context, using an LM as a major method component is extremely common. (e.g., \emph{BeamSearch} guided by LM \cite{lu-etal-2022-neurologic,beamsearch1:2021,beamsearch2:2018}). 

In practice, there is a strong relationship between the acceptability of an output and the fact that the same output receives a high probability by the LM. However, this statement must be nuanced, as perfectly acceptable samples may obtain poor probabilities in the LM testing phase. Therefore, in general, human verification cannot be avoided, but LMs are the best tools for text generation to minimize human feedback in the loop and to produce the most human-like content. 

Constraint Programming (CP) is a paradigm to compute solutions that satisfy constraints. 
CP is mainly based on filtering algorithms (also known as propagators), which remove values from variable domains that do not belong to a solution of a constraint. The association of such algorithms with global constraints is one of the main strengths of CP because they exploit the specific structure of each constraint.

Recently, a framework for generating constrained sentences in CP has been proposed \cite{bonlarron-et-al:2023} when common approaches like BeamSearch are ineffective. This constrained text generation task is viewed as a discrete combinatorial optimization problem and can be described as follows: the decision Variables represent individual words, the Domains of the variables represent the permissible lexicon, and the Constraints represent the requirements imposed on the generated text.

It focuses on satisfying strong constraints since they are the core difficulty of the problem and relies on n-grams (i.e., sequences of $n$ words of a corpus) to generate sentences of a language. Then, an LLM computes a Perplexity score for each solution, and the best-rated ones are retained. This approach solves the problem that had not yet been solved. However, it produces lots of poor sentences (more than 90\%). 

Checking the constraint first and then selecting the best solutions is suited as long as the solutions set is limited (thousands of solutions) because the scoring requires calling an LLM to evaluate each solution, which is costly ($\ge$100 ms per call). Even though this method can be reproduced for other problems, using the LLM at the very end of the method can be a drawback, mainly because sequences leading only to poorly evaluated sentences are not filtered. In this way, the LLM is not involved in controlling the combinatorial explosion. It prevents increasing the corpus size or lowering the order of n-grams, which significantly increases the number of solutions to be evaluated, and scoring and ranking millions of solutions with an LLM is impractical. For example, Bonlarron et al.'s method uses 5-grams and is challenging to use with 4-grams.

Therefore, an LLM-based constraint should be designed to efficiently filter out "poor" solutions during the solving. This task is complex for two reasons: 
\begin{enumerate}
\item The LLM probability computation depends on the elements in the sequence, so it is difficult to decompose. For a given sequence $w_0...w_n$ where $w_i$ are words, we have $P_{LLM}(w_n | w_0...w_{n-1})$. The probability is handled in major part by the attention layers \cite{vasmani-et-al:2017} that are not known for their clear interpretability. Therefore, it is unclear how to obtain from $P_{LLM}(x_n | x_0...x_{n-1})$, two sub expressions like $P_{LLM}(x_i | x_0...x_{i-1})$ and $P_{LLM}(x_j | x_{i+1}...x_{j-1})$ , where $i+j=n$. It is consequently difficult to estimate an even finer decomposition as a function of the variables (words) and to define a filtering algorithm. 
\item The LLM queries are time-consuming\footnote{https://docs.nvidia.com/nemo-framework/user-guide/latest/performance/gpt.html\#inference-performance}. This prevents systematic interaction with an LLM during the search, since it may produce millions of calls.
\end{enumerate}

To overcome these difficulties, we propose to define the \emph{NgramMarkov} constraint, a variant of the Markov constraints. 
This constraint is based on a set of n-grams and uses an LLM.

N-grams provide a natural mechanism for enforcing the linguistic constraints locally by defining the acceptable transitions between words. Specifically, two n-grams can be linked if they share at least n-1 words in common. For instance, in the sentence: "The little boy plays with a balloon", several 2-grams can be found (e.g., ``The little",``little boy", ``boy plays", ``plays with", ``with a", ``a ballon"). 

The main issue when using n-grams is the linguistic mistakes caused by local decisions. For instance, some transitions do not take into account the whole sentence or some critical words in the past. 
 
This leads to three common linguistic mistakes while chaining n-grams: 1) syntactic errors, 2) grammatical errors, and 3) lack of meaning and coherence. To illustrate, a classical pitfall when dealing with n-gram is an agreement mistake between the subject and another word related to it in the sentence. This arises when the distance (in the number of words) is greater than the size of the history (n-1, where n is the size of a given n-gram). For instance, consider the generated text   
\emph{John sometimes sends letters to himself/herself}). The decision to pick \emph{himself} instead of \emph{herself} considering the whole sentence is obvious, but both are possible while considering n-gram with $n$ lesser than $6$. Despite these limitations, where n-grams are blind to some mistakes, n-grams help to produce correct and meaningful content.

The n-grams can be seen as particular transition functions between their words. To filter n-grams, we can associate probabilities with n-grams (like the maximum likelihood estimation) and use the constraints \emph{ElementaryMarkov} \cite{pachet-roy:2011}  or \emph{MDDMarkovProcess} \cite{perez-regin:17b}. However, the filtering associated with the \emph{ElementaryMarkov} constraint is weak when used with n-grams as a transition function.
Since the text is represented as a succession of $k$ words (k-gram), each value (word) in the domain of the variables keeps its support until a particular k-gram is chosen, so there are almost no future values removed by the \emph{ElementaryMarkov} constraint.

The \emph{MDDMarkovProcess} constraint is more global than the \emph{ElementaryMarkov} constraint and the associated propagator is based on the costMDD propagator \cite{perez-regin:17} and so is more powerful. 
Unfortunately, as the name suggests, this constraint implies working with a MDD (Multi-valued Decision Diagram) that represents all the words in the sentence whose probability we want to control. While this solution may be possible with a small corpus and 5-grams, it is no longer possible with a large corpus or 4-grams. This solution is therefore not generally acceptable. Nevertheless, we will reuse the ideas of the costMDD propagator.

To deal with these issues, the propagator of the \emph{NgramMarkov} constraint is a natural extension of that of the \emph{ElementaryMarkov} constraint by incorporating the LLM distribution as a replacement for the maximum likelihood estimation of the n-grams and by adding more filtering criteria.

On the one hand, it addresses two significant issues when dealing with n-gram models: 1) sampling complexity (i.e., the lack of example to estimate the probabilities of high order n-gram, $n \geq 3$ ), and thus avoiding all smoothing and interpolating algorithms used as a workaround.
2) Sensitivity control of n-gram-based generation, to control the combinatorial explosion that occurs with small n-grams (some n-grams have a branching factor of 200). 

On the other hand, it introduces a gliding threshold to impose global regularity. This threshold guarantees that the data do not deviate locally too much from the statistical distribution of n-grams (e.g., mean or standard deviation). It also uses a "look-a-head" approach to consider only potential successions of a given n-gram over a fixed length horizon in the future. This idea, inspired by the propagator of the \emph{MDDMarkovProcess} constraint,  avoids considering minimum or maximum probabilities that cannot occur in the near future of a given n-gram.

The article is organized as follows: Sec.~\ref{sec:Prelim} provides reminders of the Large Language Model (LLM) and n-gram model to ensure a full understanding of the approach. Sec.~\ref{sec:method} introduces the \emph{NgramMarkov} constraint and explains how to obtain an LLM-powered n-gram model. Also, it then details the propagator and, in particular, the various filtering criteria that can be used to filter out n-grams. Sec.~\ref{sec:resultats} demonstrates the effectiveness of the approach by sharing several benchmarks associated to each criterion. Additionally, it compares the different strategies in terms of quantitative (e.g., number of solutions) and qualitative analysis (e.g., Perplexity scoring). Subsequently, Sec.~\ref{sec:discussion} gives some perspectives about this work, and Sec.~\ref{sec:conlusion} concludes this paper.

\section{Preliminaries}
\label{sec:Prelim}

\subsection{LM Constraints}

The \emph{ElementaryMarkov} constraint (EMC) \cite{pachet-roy:2011} is the first LM constraint.  
It was designed to address music generation in CP by leveraging Markov chains. 
An EMC is a triplet (context, continuation, prob) where prob is the likelihood of the continuation given the context. The EMC is repeated for each variable.
In a nutshell, EMC allows a CP formulation of a Markov Process \cite{morin-Quimper-markov:2014}. 
Therefore, this contribution enabled statistical guarantees in the solution based on n-gram models for music generation and other prospects. For instance, it allows the melody of a style of music in CP to be modeled (e.g., blues).

The \emph{MDDMarkovProcess} constraint \cite{perez-regin:17b} is a global LM constraint. It utilizes an MDD to ensure a lower $P_{min}$ and an upper bound $P_{max}$ of probability with respect to a given Markov Process $M$ of any solution. This constraint can be viewed as a summation constraint computed in a Cost-MDD, where the costs are logprob. 

\subsection{Language Model}
Language Models (LMs) assign probabilities to sequences of words. Various techniques have been developed to achieve this goal. The following sections briefly introduce two of the most relevant techniques. 

\subsubsection{N-gram Model}

An n-gram model \cite{jurafsky:2009,shannon-ngram:51} estimates the probabilities of sequences of words based on a sliding window of size $n$. (i.e., a context of $n$ words). This assumption is known as the Markov assumption, which states that the probability of a state can be determined solely based on the previous state. The previous state is referred to as a history $H$, and the size of the history corresponds to $n-1$ for a given n-gram size.

For instance, in a bi-gram model ($n = 2$), the computation only relies on the previous word to predict the next two words in a sequence of words. As a result, the history length is equal to one. If $Y$ is denoted as the following n-gram, the computation is as follows:

\[P_{n-gram}(Y | H) = \frac{C(Y)}{C(H)} \]
where C is the number of occurrences of the n-gram. This method of calculating probabilities is known as Maximum Likelihood Estimation (MLE) \cite{jurafsky:2009}. Consequently, MLE requires a corpus (a set of sentences) to perform this computation. This is a significant limitation because, in real-world applications, this substantially limits the use of higher-order n-grams ($n > 3$), because if the corpus is not large enough, a lot of n-grams are present in only one or two sentences and the MLE assigns them a probability of $1$ or $1/2$.

Nevertheless, despite the difficulty of gathering sufficient data to compute accurate probabilities, n-gram models are computationally efficient and cost-effective.

In general, the use of n-grams as a building block to produce text or music that satisfies some properties has been successful in CP (e.g., style modeling~\cite{pachet-roy:2011}, virtuoso melodies~\cite{pachet-roy-barbieri:2011-virtuoso}, meter~\cite{Roy-Pachet:2013-meter}, plagiarism~\cite{papadopoulos-roy-pachet:2014}, palindromes generation~\cite{papadopoulos-roy-etal:15}, poetry~\cite{perez-regin:17b}, standardized sentences~\cite{bonlarron-et-al:2023}).

\subsubsection{Large Language Model}

Large language models (LLMs) \cite{openai2023gpt4,touvron2023llama} represent the latest generation of language models, primarily relying on powerful neural network architectures. They are trained on massive amounts of data and usually utilize a specific architecture based on attention called Transformer. Despite these technical considerations, their primary goal remains to compute probabilities for sequences of words based on training data. They manipulate word embedding, and two significant parameters play a role in the generation/prediction of the text of LLM. 

\begin{itemize}
    \item Attention: This mechanism enables LLMs to track relationships between words.
    \item Position: LLMs can incorporate positional encoding, allowing them to account for the order in which words appear in a sequence.
\end{itemize}
Note that Attention and Position are two parameters that are not considered in the n-gram Model.
The primary drawback of LLMs is their computational cost. Due to their immense size and architecture, they require powerful hardware, typically clusters of GPUs, to operate effectively (several models need more than 80GB of GPU VRAM).

\section{Method}
\label{sec:method}
Our method is based on word sequence filterings using probabilities derived from the n-gram model.

Given two n-grams $N_1$ and $N_2$ of size $n$, $N_2$ is a successor of $N_1$ if the $n-1$ last words of $N_1$ are the $n-1$ first words of $N_2$.

\begin{definition}
Given 
\begin{itemize}
\item ${\cal N}$ a set of n-grams, 
\item $X=\{x_1,x_2,...,x_n\}$ a set of variables whose domain are included in ${\cal N}$,
\item $LLM$ a large language model,
\item $P_{max}$ a maximum probability.
\end{itemize}
The constraint  \texttt{NgramMarkov}$(X,{\cal N},LLM,P_{max})$ ensures that for every allowed tuple $t=(N_1,N_2,...N_{|X|})$, the n-gram $N_i$ is a successor of the n-gram $N_{i-1}$ and $t$ satisfies
\[\sum_{i=1}^{|X|} \log (P_{LLM}(N_i)) \leq \log(P_{max})\]

 where $P_{LLM}(N_i)$ is the probability value given by the LLM for the n-gram $N_i$.
\end{definition}

The constraint leverages the Markov assumption to approximate the probability computation of an LLM.
Roughly, it is a practical estimation of the sequence likelihood.

More precisely, instead of relying on MLE, the constraint utilizes probability values obtained from the LLM. Namely,
 \[ P_{n-gram}(Y | H) = \frac{C(Y)}{C(H)}, \]
 is replaced by
\[ P_{n-gram}(Y | H) \approx P_{LLM}(Y | H) =  P_{LLM}(Y).  \]

This involves querying the LLM for each n-gram and storing the returned probability in the constraint. This approach mitigates the computational overhead of direct LLM calls, which can reach up to 100 milliseconds per query.

\subsection{Filtering Criteria of N-grams}
\begin{figure}[tbp]
    \centering
\includegraphics[width=\textwidth]{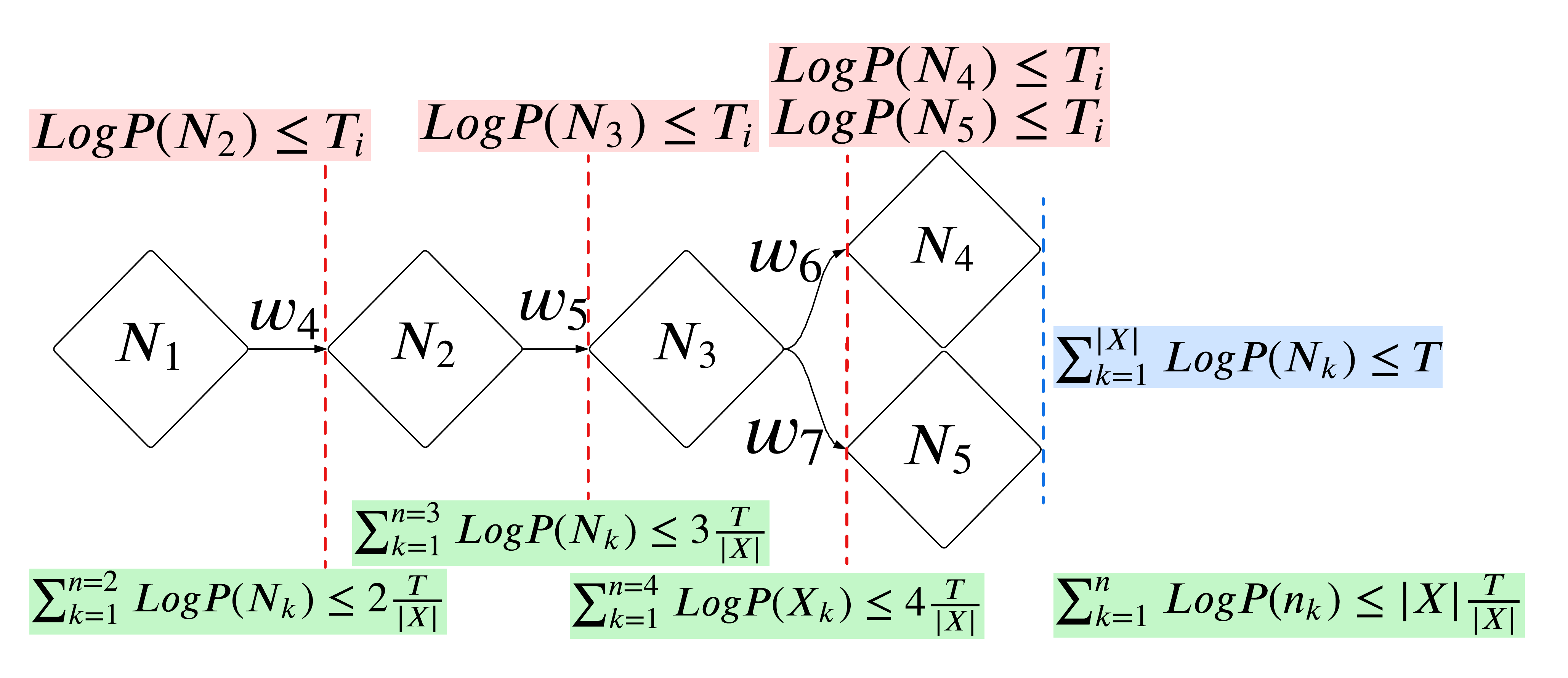}
    \caption{This figure illustrates a Markov Process in a simplified case with several 3-grams. Where each state is a 3-gram (e.g., $n_2=w_2w_3w_4$) and each transition between state is a word (e.g., $w_5$). 
    It highlights each filtering criteria of n-gram based on their LogProb: 
    (1) In \color{red}red\color{black}, the instant Threshold (renamed $T_i$, to avoid confusion with the final one $T$), for each possible word transition.
    (2) Then, in \color{blue}blue \color{black}, the final threshold, checks only the total sum of logprob is under the threshold $T$.
    (3) Next, in \color{OliveGreen}green\color{black}, the gliding threshold integrates the final threshold at each transition step, checking if the partial sum at step $k$ falls under $k * \frac{T}{|X|}$.
    Finally,  the various dashed-lines indicates where the constraints check are called}
    \label{fig:summary-criteria}
\end{figure}

In this section, several filtering criteria of n-grams are presented (See Fig. \ref{fig:summary-criteria}). 
These filtering algorithms are called only for the assigned variables. 

\subsubsection{Transition Filtering: Instant Threshold}

Consider an n-gram of size $n$, denoted as \[N_j = w_aw_b...w_e,\] along with its $k$ successors, 
\[succ(N_{j})= \{N_{s_1}=w_b...w_ew_{s_1},...,N_{s_k}=w_b...w_ew_{s_k}\}.\] 
The decision to keep $N_{s_i}$ as a potential value for the next variable is based on $\log P_{LLM}(N_{s_i})$. If the probability of chaining the current n-gram $N_j$ with the next n-gram $N_{s_i}$ falls below the given threshold $T$ (i.e., $\log P_{LLM}(N_{s_i}) \leq T$), then $N_{s_i}$ is filtered out. The n-gram $N_{s_i}$ is no longer considered for assignment to the next variable. This constitutes a local filtering criterion involving a word-by-word filtering step.

\subsubsection{Path Filtering: Final Threshold}

A global criterion involving the whole sequence $S$ can be made. Several n-grams ($N_i$) are chained during the generation to produce the final sequence.
Therefore, the probability of $S$ can be expressed as the conditional probability of the n-gram chained.
$P(S)$ can be written as \[P(N_1...N_{|X|}) = P(N_n | N_1..N_{|X|-1}) = \sum_{k=1}^{|X|} \log P(N_k). \]
Then check the following filtering criterion: \[\sum_{k=1}^{|X|} \log P_{LLM}(N_k) \leq T, \] ensures that any sequences produced have a log probability below a given threshold $T$. 
This formulation corresponds to the propagator of the \emph{MDDMarkovProcess} constraint \cite{perez-regin:17b} with $T$ replacing $P_{max}$.

\subsubsection{Prefix-based Filtering: Gliding Threshold }
To control the generation, the instant threshold and the final threshold can be fused. On the  one hand, some transitions from a local point of view may be particularly unlikely, even though these still exist in the language\footnote{Some AI-generated text detectors work by following this assumption:  if a given text contains unlikely words in several positions, it should be human-made. (see http://gltr.io)} \cite{gehrmann-etal-gltr:2019}. On the other hand, the whole sequence must have no valuation with a probability too low. Then, threshold $T$ is expressed as a function of the current number of n-grams appended in the sequence so far. It allows some very unlikely transitions to happen, but at the same time, it ensures a high overall probability. The most straightforward way to define $T$ is to divide it by the number of n-grams needed (i.e., the number of words needed in the output). The main intuition behind the gliding threshold is that it enforces a sort of regularity (i.e., a limited spread between two probability valuations) compared to the Final Threshold. 

For each value $k$ from 1 to $|X|$, there is one inequality derived from:
\[\sum_{j=1}^{k} \log P(N_j) \leq k\frac{T}{|X|}\]

This expression can be fine-tuned by taking account of the statistic summary extracted from the n-gram distribution. First, by allowing a bit of slacking that the constant $C_{slack}$ quantifies (it allows to start more often with unlikely n-gram of the start of a sentence). 

For each value $k$ from 1 to $|X|$, there is one inequality derived from:
\[\sum_{j=1}^{k} \log P(N_j) + C_{slack}  \leq k\frac{T}{|X|}\]

Then T can be defined, thanks to the Mean $\mu$ and the Standard Deviation $\sigma$ of the n-gram distribution.

For each value $k$ from 1 to $|X|$, there is one inequality derived from:
\[\sum_{k=1}^{k} \log P(N_j) + C_{slack}  \leq k(\mu - (\lambda\sigma))\]

Finally, $C_{slack}$ can be defined as one standard deviation (i.e., $C_{slack} = \sigma$), and $\lambda$ factor defines how n-grams that do not belong to the right tail of the n-gram distribution are penalized. Its range is $[0,2]$.

\subsubsection{Look-a-head Filtering}
 Chaining between n-grams is strongly constrained because there must be $n-1$ words in common. Also, the number of n-grams that can be reached in the near future from a single given n-gram is quite small compared to the overall n-gram set. As a result, the probabilities for future n-grams are also quite specific. It is therefore advisable to use a filtering algorithm for the n-grams possible for the $k^{th}$ word generated, which can be seen as a shaving algorithm: for each n-gram and for a given depth $p$ (where the depth $p$ defined the horizon in the future), only the n-grams that can be reached up to $p$ are considered in the previous filterings. This enumeration of n-grams could be stored in an MDD of depth $p$.
 
 This makes a significant difference to the algorithm that considers all possible n-grams simultaneously, as the n-grams reached may have very different probabilities, and the minimum or maximum probability calculated for each potential future word is no longer necessarily related to the n-gram constraint.

 Thus, an n-gram is filtered out if none of its future transitions (in $p$ steps) may lead to a high enough probability sequence.

For each value $k$ from $i$ to $i+p$, there is one inequality derived from:

  \[ \sum_{j=i}^{i+p} \log P(N_j) \leq (k-i+1)\frac{T}{|X|}  \]

Then $\frac{T}{|X|}$ is replaced in the same way as for the gliding criterion, i.e. $(\mu - (\lambda \sigma))$.

The relationship between two n-grams extends as long as the minimum distance between their words is less than $n$. If the distance between them exceeds $n$, any n-gram can follow, regardless of its linguistic relation to the first n-gram. 
This reasoning can be illustrated with the same example from the introduction with the sentence: \emph{John sometimes sends letters to himself/herself}. With 5-gram, it is possible to use \emph{herself}. It is not possible with n-gram with $n \geq 6$. Therefore the length of the horizon is not greater than $n$.

\section{Results}
\label{sec:resultats}

\subsection{Experimental Conditions}
The approach described in Sec.~\ref{sec:method} is implemented in Java~17. 
The code is available upon request. 
\paragraph{Application:}The propagator is evaluated on a standardized sentences generation task \cite{bonlarron-et-al:2023}. 
\paragraph{Generation:}The generation experiments were performed on a machine using an Intel(R) Xeon(R)~W-2175 CPU~@~2.50GHz with 256 GB of RAM and running under Ubuntu 18.04. 

\paragraph{Inference:}The LLM inference experiments were performed on a machine using an AMD EPYC 7313 16-Core CPU~@~3GHz with 512 GB of RAM and an A100 GPU running under Ubuntu 20.04.6 LTS.

\paragraph{Evaluation:}
The scoring of the solution set is performed by the LLM used by Bonlarron et al.\footnote{https://huggingface.co/asi/gpt-fr-cased-base} (See \cite{simoulin:2021} for more information).

\begin{figure}[tbp]
    \begin{subfigure}{0.49\textwidth}
    \centering
    \includesvg[width=\textwidth]{ima/exp/4gram-scored.small-gpt-plot.svg}
    \caption{$\mu=-5.26$ , $\sigma=1.43$}
    \label{fig:4g-dist}
    \end{subfigure}
    \hfill
    \begin{subfigure}{0.49\textwidth}
    \includesvg[width=\textwidth]{ima/exp/5gram-scored.small-gpt-plot.svg}
    \caption{$\mu=-4.76$ , $\sigma=1.24$}
    \label{fig:5g-dist}
    \end{subfigure}
    
    \begin{subfigure}{0.49\textwidth}
    \centering
    \includegraphics[width=\textwidth]{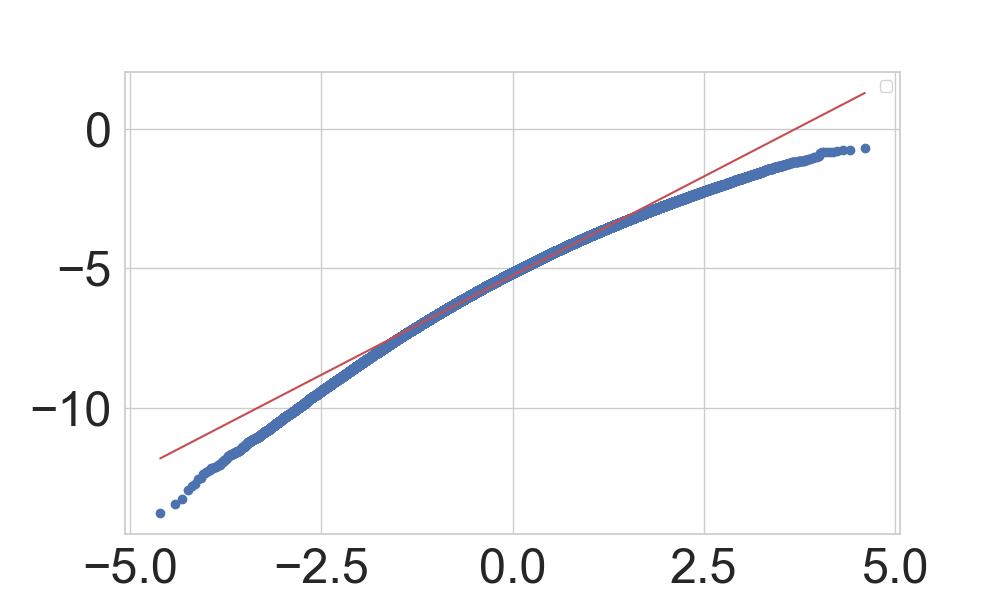}
    \caption{QQplot between (a) and normal law: left-skewed data}
    \label{fig:qq-4g}
    \end{subfigure}
    \hfill
    \begin{subfigure}{0.49\textwidth}
    \includegraphics[width=\textwidth]{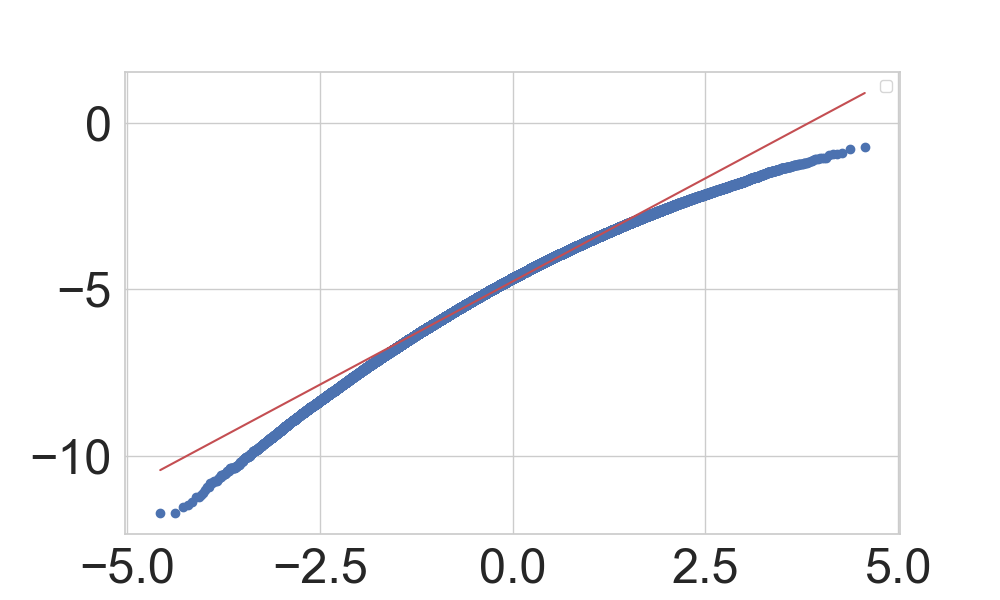}
    \caption{QQplot between (b) and normal law: left-skewed data}
    \label{fig:qq-5g}
    \end{subfigure}

    \caption{This figure draws the distribution of 4-grams (a) and 5-grams (b) extracted from french books binded with log-prob computed from a light french GPT model$^3$ (a,b) and their associated Quantile-Quartile plot (QQplot) respectively (a,c) and (b,d).\label{distrib+qq} }
    
\end{figure}

\subsection{Achieving around 10ms N-Gram Scoring}
Even though it is offline computation, scoring millions of n-grams can be extremely long. Two ways can be used to reduce as much as possible that time: (1) choosing the lightest possible model (i.e., train on a small corpus). 2) using LLM performance optimization techniques.

Consequently, the model used to score the different n-grams is the lightest French GPT-2 that can be found on Hugging-face Library \cite{hugginface:2020}\footnote{https://huggingface.co/ClassCat/gpt2-base-french}. Then, a series of optimizations are done before loading and querying it: (i) Model Quantization \cite{Jacob_2018_CVPR} (Q4)
(ii) betterTransformers (iii) fast-attention (Flash Attention \cite{dao2022flashattention}).
It takes, on average, 10 ms to score a n-gram and scoring all 5-grams took less than a day. All the French 4-grams took roughly the same. Both n-grams distributions can be seen in Fig. \ref{distrib+qq}

\subsection{Statistical Analysis}
Once the n-grams extracted from the books have been associated with a probability value calculated by an LLM, the threshold $T$ used as filtering criteria during generation must be defined.

$T$ could be found by trials and errors concerning an expected number of solutions or run time. (e.g., by taking an huge $T$ value and decreasing it each time).
However, this method is inaccurate and slow.

It is more interesting to define $T$ from the resulting distribution  of the probabilities of the n-grams (see Fig. \ref{fig:4g-dist} and Fig. \ref{fig:5g-dist}). 
The two distributions though similar to normal distribution are not. The associated quantile-quantile (QQ) plot, gives an intuition on why both fail normality test. In fact, both distributions are left-skewed (see Fig. \ref{fig:qq-4g} and Fig. \ref{fig:qq-5g}). A left-skewed QQ-plot indicates that the distribution is not symmetrical, with a longer tail extending towards the lower end of the range. This is illustrated by the concave curvature of the QQ-plot curve relative to the reference line. The imbalance between the mean and median, where the mean is pulled towards the left, further highlights the left-skewed nature of the distribution.    
Even though, the distributions are not normal, their mean and standard deviation, remains relevant values to describes them.
Therefore, they can still be used to define some interesting filtering ranges. 

In this paper, the threshold $T$ is simply defined as a linear combination of the mean ($\mu$) and the standard deviation ($\sigma$) of the n-gram distributions.

\begin{table}[btp]
    \centering
    \resizebox{\textwidth}{!}{  
    \begin{tabular}{cccccc}\toprule
    
       $n$  & nodes & arcs & mem (GB) & time (s) & sols   \\
       \midrule
       4  & $\gg$ 88,837,214 &  $\gg$ nodes    & $\gg$ 240& $\gg$2600  & OOM      \\
       5  & 14,778 & 18,463 & 3 & 68& 5,052 \\
        \bottomrule
    \end{tabular}
    }
    \caption{Number of arcs, nodes, solutions, gigabytes (GB), and seconds (s) for computing MDDMNREAD in 4-grams and 5-grams.
    Same model and datasets from Bonlarron et al.}
    \label{tab:vanilla}
    \centering
    \hspace{2cm}
    \resizebox{\textwidth}{!}{
    \begin{tabular}{lllllll}
     \toprule
         & \multicolumn{2}{c}{$T=\mu$} & \multicolumn{2}{c}{$T=\mu-\sigma$} & \multicolumn{2}{c}{$T=\mu-1.5\sigma$}  \\
       \cmidrule(lr){2-3}\cmidrule(lr){4-5} \cmidrule(lr){6-7}
        $n$  & sols & time (s) & sols & time (s) & sols & time (s) \\
       \midrule
       4  & $1.67e^7$ & 1440   & 1649 & 21s  & 1 & 2s\\
       5  & 1287 & 26 & 2 & 5 & 0& / \\
        \hline
    \end{tabular}
    }
    \caption{Instant Threshold :  $Log(P(N_k)) \leq T$}
    \label{tab:instant}

 \hspace{2cm}
 
    \centering
    \resizebox{\textwidth}{!}{ 
    \begin{tabular}{lcccccccccc}\toprule
         & \multicolumn{2}{c}{$\lambda=1$} &  \multicolumn{2}{c}{$\lambda=1.25$}& \multicolumn{2}{c}{$\lambda=1.5$} & \multicolumn{2}{c}{$\lambda=1.75$} & \multicolumn{2}{c}{$\lambda=2$}   \\
       \cmidrule(lr){2-3} \cmidrule(lr){4-5} \cmidrule(lr){6-7} \cmidrule(lr){8-9} \cmidrule(lr){10-11}
       n  & sols & time (s) &sols & time (s) & sols & time (s) & sols & time (s) & sols & time (s) \\
         \midrule
       4  & OOM & / & OOM & / & 56652 & 330 & 894 & 11 & 4 & 3\\
       5  & 1680 & 50 & 474& 17& 6 & 9 & 2 & 5 & 0 & / \\
        \bottomrule
    \end{tabular}
    }
    \caption{Gliding Treshold: $T = \mu - \lambda \sigma$. }
    \label{tab:gliding}
    \centering
\hspace{2cm}

 \resizebox{\textwidth}{!}{ 
    \begin{tabular}{ccccccccc}
     \toprule
         & \multicolumn{2}{c}{$\lambda=1$} & \multicolumn{2}{c}{$\lambda=1.25$} &  \multicolumn{2}{c}{$\lambda=1.5$} & \multicolumn{2}{c}{$\lambda=1.75$} \\
         & \multicolumn{2}{c}{$p=n-1$} & \multicolumn{2}{c}{$p=n-1$} & \multicolumn{2}{c}{$p=n-1$} & \multicolumn{2}{c}{$p=n-1$}  \\
       \cmidrule(lr){2-3} \cmidrule(lr){4-5} \cmidrule(lr){6-7} \cmidrule(lr){8-9}
        $n$  & sols & t & sols & t & sols & t & sols & t \\
       \midrule
       4  & 954140  & 9791 & 54884 &576  & 2607  & 46 & 4 & 10   \\
       5  & 162  & 11 & 25 & 5 & 0 & / & 0 & /\\
        \bottomrule
    \end{tabular}
    }
    \hspace{1cm}
    \caption{Gliding Threshold and Look-ahead: $T = \mu - \lambda \sigma$ and $p$ defines the horizon length. The horizon is computed in a DFS fashion with nested loops. Look-ahead is checked before any other criteria.} 
    \label{tab:gliding+future}

\end{table}

\begin{table}[tbp]
   \centering
    \resizebox{\textwidth}{!}{ 
    \begin{tabular}{llr}
    \toprule
    french generated sentences & DeepL translation & PPL \\
    \midrule
Elle avait trouvé un moyen de se rendre compte de la vérité. &She had found a way to realize the truth. &13 \\
Elle flottait dans le ciel au-dessus de la maison de madame. &She floated in the sky above Madame's house. &15 \\
Nous avions besoin de savoir si tu avais besoin de mon aide. &We needed to know if you needed my help. &15 \\
Puisque je viens de te demander si tu as besoin de vacances. &Since I just asked you if you needed a vacation. &16 \\
Il devait trouver un moyen de se rendre à la maison de jeux. &He needed to find a way to get to the playhouse. &19 \\
Même si vous avez déjà entendu parler de la petite sorcière. &Even if you've already heard of the little witch. &20 \\
On allait trouver un moyen de se rendre à la fête de demain. &We were going to find a way to get to tomorrow's party.  &22 \\

Le roi était en train de passer au-dessus de la mer immense. &The king was passing over the immense sea.  &24 \\
Il repoussa la jeune femme à la maison de la rue principale. &He pushed the young woman back to the house on the main  street. &40 \\

\bottomrule
    \end{tabular}
    }
    \caption{Cherry-picked French sentences in 4-grams, a DeepL$^5$ translation is provided with each sentence.}
    
    \label{tab:cherry}

    \centering
    \resizebox{\textwidth}{!}{ %
    \begin{tabular}{llr}
    \toprule
    french generated sentences & DeepL translation & PPL \\
    \midrule
Ils sont un peu plus tard dans la nuit de samedi à dimanche& They are a little later in the night from Saturday to Sunday &9 \\
Elle est un peu plus tard dans la nuit de samedi à dimanche& She is a little later in the night from Saturday to Sunday &10 \\
Il reste un peu plus tard dans la nuit de samedi à dimanche& He stays a little later in the night from Saturday to Sunday &11 \\
Agissons comme si nous étions à la fin de la dernière étape& Let's act as if we were at the end of the last stage &11 \\
Je ne connais pas le nom de la première partie de la maison& I don't know the name of the first part of the house &11 \\
Tu étais en train de passer au-dessus de la première maison&You were passing over the first house &13 \\
Elle se trouve dans le ciel au-dessus de la première maison& It's in the sky above the first house &13 \\
Agissons comme si nous étions à la fin de la dernière pluie&Let's act like we're at the end of the last rain &14 \\
Il arrive un jour où je me rends compte de la vie elle-même& There comes a day when I realize life itself   &15 \\

\bottomrule
    \end{tabular}
    }
    \caption{Some best ranked French sentences in 4-grams, a DeepL$^5$ translation is provided with each sentence.}
    \label{tab:sentences}
\end{table}

\footnotetext{https://www.deepl.com/translator} 

\subsection{Performance Analysis}

For the sake of clarity, Bonlarron et al.'s approach is renamed the "vanilla model".
First, the results obtained with the vanilla model are reproduced in Tab.\ref{tab:vanilla}.  
5,052 solutions are generated for 5-grams, and it is not possible to deal with 4-grams.
Tables~\ref{tab:instant},~\ref{tab:gliding},~\ref{tab:gliding+future} give the results obtained with different filtering criteria of the \emph{NgramMarkov} constraint.

All of the criteria associated with the new constraint eliminate many sentences from the vanilla model, which is an expected result. The Instant Threshold and Look-ahead criteria successfully compute solutions set with 4-grams. The Gliding Threshold criterion does not have enough memory when $\lambda~\leq~1.25$. The Instant Threshold criterion applies a selective filtering. By ensuring that all n-grams in a solution meet a minimum probability threshold, many solutions are filtered.

The Gliding Threshold criterion offers a more nuanced filtering approach. Although it filters less than the Instant Threshold, it captures more information about word sequences. 
It does not filter sufficiently for $\lambda \le 1.25$ and fails to prevent an out-of-memory (OOM) error. There is no noticeable overhead in computing solutions compared with Look-ahead. In fact, to obtain 56652 solutions in 4-grams with $\lambda =1.5$ for Tab. \ref{tab:gliding}, it takes 330  seconds, whereas to obtain an equivalent number of solutions with look-ahead with $\lambda=1.25$ it takes 576 seconds. This can be explained by the fact that look-ahead explores for each n-gram its future successors on the $p$ horizon in Tab. \ref{tab:gliding+future}. However, it allows to check strong properties on the future of n-grams, resulting in strong filtering even on permissive values like $\lambda=1$.

The Look-ahead criterion is able to compute the solution set for the most interesting values of $\lambda$ (i.e. $\lambda \geq 1$). With 5-grams, it tends to remove too many solutions (i.e. $\lambda > 1$). However, for $\lambda =1$, it deletes many solutions but retains a large proportion of the best ones.

The vanilla model selects sentences using the perplexity computation. It is, therefore, essential to compare the perplexity-scored sentences of the different approaches.

\begin{figure}[tbp]
    \centering
\includesvg[width=\textwidth]{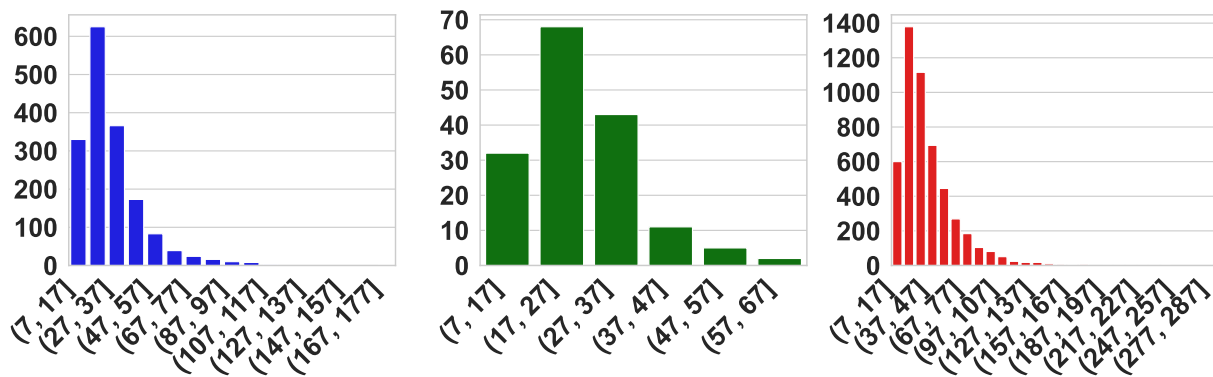}
    \caption{Comparisons of PPL distribution of sentences produced in 5-grams between: gliding threshold (blue) for $\lambda=1$, look-ahead threshold for $\lambda=1$ (green) and vanilla model (red).}
    \label{fig:distrib_ppl}
\end{figure}

Fig. \ref{fig:distrib_ppl} compares the PPL distribution of the generated sentences when using 5-grams for the Gliding Threshold criterion, the Look-ahead criterion, and the vanilla model.
The perplexity interval of the vanilla model is $(7,292]$, and the mean is $38$. 
The Gliding Threshold criterion
 eliminates many outliers, but the statistical range remains large $(7,177]$ with a mean equals to $29$, whereas for Look-ahead criterion the statistical range is greatly reduced $(7,66]$ with a mean equals to $26$.
 A maximum value for the perplexity in order to help the sentence selection seems to be $27$.
With this value, the vanilla model keeps $2,000$ sentences, the Gliding Threshold criterion $1,000$, and the Look-ahead criterion $100$. Depending on the number of required sentences, a particular criterion and its associated lambda value can be chosen.

\subsection{Limitations}
\paragraph{In 5-grams:}
The approach makes the sentence selection parts in the solutions set more manageable despite some good sentences being lost. The loss of some good solutions in the process is a clear drawback. Nevertheless, it empirically removes many poor solutions, so the approach remains useful.

\paragraph{In 4-grams:}
The novel propagator is essential since the vanilla model made 4-grams computation intractable. However, the mistakes that occur occasionally in 5-grams appear regularly in 4-grams. As a result, the proportion of poor solutions is significantly increased.

\section{Discussion}
\label{sec:discussion}

\subsection{Perplexity Ranking}

During the selection of sentences generated from 4-grams, a new difficulty appears.
Tab. \ref{tab:sentences} shows a short sample of the best-scored 4-gram solutions. 
They are correct French sentences but are often not self-contained. The three first illustrated it somehow.
\emph{They are a little later in the night from Saturday to Sunday}.
The verb \emph{be, i.e., are} does not sound appropriate here, even though it only makes sentences weird but not incorrect. The sentence would not sound strange in a particular context (e.g., "they" refers to trains). Whereas, the third sentence: \emph{He stays a little later in the night from Saturday to Sunday} is not the best sentence that one could imagine, but it is context-free. The problem is that the LLM slightly prefers the first sentence (PPL=9) (not context-free) instead of this one (context-free) (PPL=11). 
This subtle difference is, in practice, substantially increased in other cases. 
Thus, the sentence selection induced by 4-grams is more intricate, although good sentences can still be found semi-manually (see~Tab. \ref{tab:cherry}).

\subsection{Related Works}
Several constraints based on statistical intelligence, such as the \emph{Spread}~\cite{peasant-regin:2005-spread} or the \emph{Deviation}~\cite{Schaus-regin:2007-deviation} constraints have been designed.

Our work fits into this theme because we are looking for well-balanced solutions in the sense of a statistical distribution.
An MDD cannot be used to represent n-grams finely, as it would be too big, but we can ask ourselves whether we could use an automaton to model our problem using the \emph{Cost-Regular Constraint}~\cite{cost-regular:2006} that uses automaton instead of MDD. Recent work shows that it is possible to approximate a probabilistic model (like a RNN) in a weighted finite automaton \cite{suresh-approximating:2021}. This shows that it would certainly be possible to do the same with an LLM.
The problem is that the \emph{Cost-Regular Constraint} works on a Directed Acyclic Graph, which is quite close in size to an MDD.

\subsection{Perspectives}
\subsubsection{Calls Parsimoniously the LLM}
The possibility of spending some time querying LLM during the search becomes realistic with the new constraint. 
The Markov assumption (i.e., $P_{n-gram}$) used to approximate sequence probability, while useful, is not without its limitations and can lead to errors. 
Calling with parsimony, an LLM could eventually correct part of the mistakes made with the n-gram model. In detail, in the ``false positive" filtering where $P_{n-gram}$ gives an acceptable probability, a computation involving the whole sequence with an LLM may detect a mistake regarding long dependency between words.
However, it is difficult to make a decision based on an LLM forecast.
\subsubsection{Creativity Assistant}
The various works by Pachet and Roy have one thing in common. It is all about interactivity with the user, especially in the context of musical composition. One of their aims was for the system to stimulate the creativity of a musician. Therefore, fast and accurate results are expected.
Looking at the results, particularly the number of sentences generated for certain criteria, the constraint filters out a huge number of solutions, which can be a little disappointing when considering the original aim of Bonlarron et al. (to obtain new sentences for application in vision research). However, other applications can be considered, such as the generation of song lyrics. In that case, the proposed propagator has excellent potential in interactive mode, where a user can obtain a few solutions quickly (i.e., almost in real-time).
\section{Conclusion}
The \emph{NgramMarkov} constraint has been presented. It surrogates a Large Language Model thanks to an n-gram model.
The filtering criteria enforce a limitation over the probability computation of the n-gram of the sentences generated. The Gliding Threshold and Look-ahead criteria demonstrate their effectiveness by computing solutions set with 4-grams and 5-grams.
The experimental analysis of the sentences generated shows that the constraint positively affects the sentence selection task performed by the perplexity ranking. Reducing the number of candidate sentences and deleting outlier values (i.e., poorly scored sentences) enforces regularity concerning the mean and the standard deviation of the n-gram distribution.
A better control of n-gram-based generation in CP is now possible because it significantly avoids scaling issues caused by the corpus or the n-gram size. 
\label{sec:conlusion}

\section*{Acknowledgments}
We are grateful to Nadjib Lazaar from LIRMM (University of Montpellier, CNRS, France) for his insightful question on how to incorporate learnings from the solution set into future iterations of the work.

This work has been supported by the French government, through the 3IA C\^ote d'Azur Investments in the
Future project managed by the National Research Agency (ANR) with the reference number ANR-19-P3IA-0002.

\bibliographystyle{unsrt}

\bibliography{ijcai23}

\end{document}